# Break The Spell of Total Correlation in $\beta$−TCVAE

Zihao Chen    Wenyong Wang    Sai Zou

**Abstract**

*In the absence of artificial labels, the independent and dependent features in the data are cluttered. How to construct the inductive biases of the model to flexibly divide and effectively contain features with different complexity is the main focal point of unsupervised disentangled representation learning. This paper proposes a new iterative decomposition path of total correlation and explains the disentangled representation ability of VAE from the perspective of model capacity allocation. The newly developed objective function combines latent variable dimensions into joint distribution while relieving the independence constraints of marginal distributions in combination, leading to latent variables with a more manipulable prior distribution. The novel model enables VAE to adjust the parameter capacity to divide dependent and independent data features flexibly. Experimental results on various datasets show an interesting relevance between model capacity and the latent variable grouping size, called the "V"-shaped best ELBO trajectory. Additionally, we empirically demonstrate that the proposed method obtains better disentangling performance with reasonable parameter capacity allocation.*

## 1. Introduction

Deep learning has made breakthroughs in unsupervised representation learning[1, 2], speech recognition [3-5], face recognition [6-8], target monitoring [9-12], and natural language processing [13-15]. These developments are largely due to the proposal on latent representation theory. Disentanglement representation learning is a method of latent representation that aims to separate data features for human understanding.

Currently, many representation learning models are based on the VAE proposed by Kingma [1]. Higgins et al [16] proposed that a certain degree of penalty coefficient can be imposed on the KL term of VAE to strengthen the independence constraints of the posterior approximate distribution. R. T. Chen and Kumar et al [2, 17-19] focused on penalizing the total correlation of the latent variables $D_{KL}(q(z)||\prod_i q(z_i))$, reducing the mutual information between the latent variables to achieve a good disentangling performance. Unfortunately, VAE's generation process needs to make assumptions about the distribution space of latent variables, which will cause the approximation error [24-27]. Locatello[20] pointed out that a purely unsupervised representation learning model cannot obtain the desired disentangling performance without giving an inductive bias. Based on Locatello's work, many representation learning models established on weakly supervised learning have appeared [21-23].

Generally, we assume that data features can be divided into independent features and dependent features[16]. Intuitively, it is also necessary to correctly allocate parameter capacity in parametric neural networks to make the disentangled learning for certain data more effective. If we apply flat disentangling constraints to all dimensions of the latent variable using total correlation, like $\beta$−TCVAE. When the total parameter capacity is low, a one-dimensional latent variable may be unable to learn "over-complex" dependent features, resulting in the overfitting of independent features, in which the dependent features are mistaken for independent features and disentangled between dimensions. When the parameter capacity is high, the one-dimensional variable may blend multiple independent features into it. Because the parameter capacity space is infinite, it is difficult to select the appropriate parameter capacity so that the dependent features can just be contained in a one-dimensional variable. Therefore how to allocate the parameter capacity of learning independent and dependent features correctly becomes a challenge of unsupervised disentangled representation.

By decomposing the total correlation term in the $\beta$−TCVAE, we obtain the iterative link of the total correlation between the joint distribution and the marginal distribution. This gives us the motivation to manipulate the granularity of disentangling constraints in latent variables, to adjust the allocation of model's parameter capacity. Moreover, the variability of disentangling constraints in latent variables makes the new model with dynamic prior distribution. Compared with several common VAEs, The new model is improved in representation learning and disentangling ability through the adaptation of flexible prior distribution to data features with different complexity.

"Break the spell" in the title is a pun: the total correlation decomposition process is like treating it as a word and then breaking it down into letters. We take the first letter of each word in the title to form the new VAE name $\beta-STCVAE$.

We trained 100000+ $\beta-STCVAE$ models on 5 datasets (mnist, shape3D, dSprite, celebA, cars3D) and found the "V"-shaped best ELBO trajectory law between model capacity and the latent variable grouping coefficient. The experimental findings demonstrate that allocating model parameter capacity in most situations is not best addressed by implementing disentangling constraints on the marginal distribution of latent variables, like $\beta-TCVAE$.

The major contributions of this paper are as follows:
- We propose a new total correlation decomposition path and explore the cause of $\beta-TCVAE$ disentangling performance by gradually breaking down the total correlation.
- Based on the total correlation decomposition, a dynamic prior distribution VAE model was proposed. $\beta-STCVAE$ can manipulate the parameter capacity to learn dependent and independent data features. Experiments on numerous datasets show that the proposed model is superior in terms of representation learning and disentangling capabilities.
- Experiments also prove that reasonable allocation of model parameter capacity can flexibly control $\beta-STCVAE$'s disentanglement learning ability to micro and macro features. Based on this, some experimental patterns are summarized on parameter space.

## 2. Related Works

### 2.1. Representation Learning

The form of inductive biases to the latent variable distribution z is of vital importance in representation learning. The literature[2, 16-19, 28, 29] has suggested adding regular inductive bias to the objective optimization function, thus imposing constraints on the distribution of latent variables *z*. Higgins et al. [20] proposed $\beta$VAE that imposes a strength on the ELBO's KL divergence term, encouraging the independence of each dimension in the *z*-space. However, the constrained potential bottleneck loses feature information when the data features pass through the encoder. The losing feature information effect reduces the representation learning ability of the model, making it difficult to achieve the best trade-off between reconstruction and feature disentangling. Some studies have proposed the decomposition of ELBO: FactorVAE [18] and $\beta-TCVAE$ [2] proposed an encouraged posterior distribution that satisfied the factorial constraints to improve the disentangling ability of the model. Burgess [29] explained VAE from the perspective of information theory and claimed that gradually increasing the information capacity of latent variables can better balance the disentangling and reconstruction effects of the model. Joint-VAE [28] proposed a solution to the disentangling problem of discrete variables by extending the prior distribution type. DIP-VAE [17] achieved the disentangling performance by constraining the moments of the latent variable distribution.

There are also works based on a prior structured model, such as the Hierarchical depth ladder network [31-33], Data grouping disentanglement [34-39], and physical space combination [40, 41]. The disentangling based on data grouping mainly shares implicit features through artificial marking and divides the style features under the shared features. For example, ML_VAE [34] proposed an approach that divided latent features into content and style. Weakly supervised learning artificially selects the training samples with different styles of the same content, and VAE is then leveraged to disentangle distinct delicate styles in the same macro content. The limitation of this approach is that the VAE model cannot adjust the learning interest when faced with real data that independent and dependent features are mixed: VAE needs to use artificial labels to converge the field where it is required to disentangle representations.

HFVAE [33] was represented as a hierarchical model, and an iterative hierarchical model was obtained by decomposing the prior distribution terms of the ELBO. Coincidentally, the objective function proposed in this paper is similar to that of HFVAE [33]. The main difference is that HFVAE achieved the feature disentangling effect in style granularity by controlling the disentangling strength of different latent variable levels. In comparison, $\beta-STCVAE$ directly decomposes the total correlation term, discards the disentangling constraints term of the sub-variable in HFVAE, and manipulates the allocation of parameter capacity by changing the grouping size of latent variables.

### 2.2. Vae Approximation Error

VAE approximate error analysis is primarily based on the empirical direction of the model [42-44]. The approximate error of VAE is generally caused by (1. The mismatch between the posterior and the prior distribution; (2. the problem of posterior collapse [45, 46].

Reducing the difference between the true posterior distribution and the prior is one of the main challenges. Different approaches have been proposed, such as using importance sampling to get tighter boundaries [47]; hierarchical multi-scale framework VAE; complex and scalable approximate posterior distribution [48, 49]. Most of the insightful points focus on more complex target posterior distributions, leading to an increase in the complexity of the model and the computational burden of model training [27]. This paper provides a new analysis method: the parameter capacity allocation perspective. The objective function of β−STCVAE changes the prior distribution of latent variables by combining variables of different sizes, thus flexibly controlling the learning ability of the model to different data features.

### 2.3. Total Correlation Estimation

Statistical independence is an important indicator referred to in machine learning [52, 53], statistics [50, 51], and bioinformatics [54, 55] for measuring the degree of correlation between independent variables. It appears as a common regularization term in representation learning and deep learning. Mutual information is an information measure commonly used in information theory that represents the degree of randomness reduction of another independent variable when one independent variable is known. However, mutual information can only be leveraged to measure a pair of independent variables.

Total correlation (TC) has played an important role in representation disentanglement learning for measuring the overall correlation among multi-dimensional variables [56]. However, calculating the TC value is a major challenge because the true distribution in representation learning is unknown and can only be estimated through sampling. Cheng et al [57] decomposed TC and used mutual information bound to estimate the TC value. However, inaccuracies were found in Cheng's work, which we corrected and proposed a new TC decomposition path in this paper.

## 3. VAE with Flexible Disentanglement

### 3.1. A new path of total correlation decomposition

β-TCVAE [2], FactorVAE [18], and DIP-VAE [17] obtain a more accurate penalty term by decomposing $E_{p(x)}[KL(q(z|x)||p(z)]$ to achieve a trade-off of disentanglement and reconstruction.

β-TCVAE decomposes $E_{p(x)}[KL(q(z|x)||p(z)]$ to get the following objective:
$$E_{p(x)}[KL(q(z|x)||p(z)] = D_{KL}(q(z,x_n)||q(z)p(x_n)) \\ + D_{KL}(q(z)||\prod_i q(z_i)) \\ + \sum_i D_{KL}(q(z_i)||p(z_i)).$$
(1)

It is generally believed that the disentangling ability of β-TCVAE [2] mainly comes from two points in objective (1): 1). The mutual information between the latent variable and the data variable represented as $I(x;z)$ in the first item; 2). The independence between latent variables is represented as $TC(z)$ in the second item. The third term constrains the difference between latent variable distribution and prior distribution.

This paper mainly studies the second item. Cheng [57] proposed to use of mutual information to estimate the total correlation, which is similar to the decomposition process of the total correlation proposed in this paper. Cheng proposed the total correlation decomposition from two paths, we name them: Top-Down Tree-like decomposition and Line-Like decomposition. The proposed decomposition path in this paper is called the Bottom-Up Tree-like path. However, their work has some inaccuracies, and the derivation results are inconsistent with ours. (See supplementary material A.4 for comparison and analysis of the three decomposition paths. supplementary materials A.1, A.2 are our revisions to their proposed paths derivation)

The following is a brief introduction to the decomposition path we proposed. The complete decomposition process is put in the supplementary materials A.3. Assuming that $z$ has $n$ dimensions, then:

$$TC(z) = D_{KL}(q(z)||\prod_k q(z_k)) =$$
$$\begin{cases} \sum_{cp(i,j)} I(z_i;z_j)|_{q(z)=q(z_i,z_j)\cdot q(z_{-i,j})} + D_{KL}(q(z)||\prod_{cp(i,j)} q(z_i,z_j)) &, 1^* \\ \sum_{cp(i,j)} I(z_i;z_j)|_{q(z)=q(z_i,z_j)\cdot q(z_{-i,j})} + D_{KL}(q(z)||\prod_{cp(i,j)} q(z_i,z_j) \cdot q(z_r)) &, 2^* \end{cases}$$
(2)

Where $1^*$ represents $n$ is even, $2^*$ represents $n$ is odd. $z_i$ represents the $i$th dimension's latent variable in the multidimensional latent variable $z$. $cp(i,j)$ represents a mechanism for selecting a pair of latent variables $z_i, z_j$ without replacement. $cp(i,j)$ in the summation symbol is consistent with the result of $cp(i,j)$ in the cumulative multiplication symbol. $z_r$ represents the final remaining latent variable after selecting several times without replacement under the $cp(i,j)$ of odd case. $p(z_{-i,j})$ represents the joint distribution probability of all variables except the $z_i, z_j$ latent variable. $I(z_i;z_j)|_{q(z)=q(z_i,z_j)\cdot q(z_{-i,j})}$ represents calculating the mutual information of $z_i$ and $z_j$ using importance sampling under the target distribution where $q(z_i, z_j)$ and $q(z_{-ij})$ are independent of each other.

Explanations of each item after total correlation $D_{KL}(q(z)||\prod_k q(z_k))$ decomposition are as follows:

The first term $\sum_{cp(i,j)} I(z_i;z_j)|_{q(z)=q(z_i,z_j)\cdot q(z_{-i,j})}$ is the "mutual information summation" term, which represents the accumulation of mutual information combinations of all pairs $z_i, z_j$ of latent variables selected by $cp(i,j)$.

The second term $D_{KL}(q(z)||\prod_{cp(i,j)} q(z_i,z_j))$ under even case, so is $D_{KL}(q(z)||\prod_{cp(i,j)} q(z_i,z_j) \cdot q(z_r))$ under odd case, called the "joint distribution total correlation" term. During derivation, we discovered that the target distribution of the importance sampling in the first term $I(z_i;z_j)$ satisfy $q(z_i,z_j)$ and $q(z_{-i,j})$ are independent of each other. The second term can also be regarded as the accuracy constraint for calculating the mutual information

summation term, avoiding sampling bias by ensuring that the proposed distribution $q(z_i, z_j)$ is not too far from the target distribution during importance sampling.

Then we set $TC_{joint\_1}(z)$ as:

$$TC_{joint\_1}(z)$$
$$= \begin{cases} D_{KL}(q(z) || \prod_{cp(i,j)} q(z_i, z_j)) & , n \text{ is even} \\ D_{KL}(q(z) || \prod_{cp(i,j)} q(z_i, z_j) \cdot q(z_r)) & , n \text{ is odd} \end{cases}.$$
(3)

We replace $\sum_{cp(i,j)} I(z_i; z_j)|_{q(z)=q(z_i,z_j)\cdot q(z_{-i,j})}$ with $MU_{joint\_1}(z)$, Then $TC_{joint\_1}(z) = TC(z) - MU_{joint\_1}(z)$.

From objective (2), knowing $MU_{joint\_1}(z)$ is always positive, defined as the accumulation of mutual information. Thus, we have $TC_{joint\_1}(z) \leq TC(z)$. This inequality brings us two interesting inferences:

**Inference 1**: The total correlation of joint distributions must be less than or equal to the total correlation of marginal distributions that compose the joint distributions.

**Inference 2**: We can split and refine the constraint granularity of $TC(z)$. $MU_{joint\_1}(z)$ constrains the independence of marginal distributions within joint distributions, while $TC_{joint\_1}(z)$ constrains the independence between joint distributions.

Furthermore, the $TC(z)$ decomposition has an iterative relationship and can continue by treating $TC_{joint\_1}(z)$ as a new $TC(z)$. All joint distribution $q(z_i, z_j)$ selected by $cp(i,j)$ in the $TC_{joint\_1}(z)$ are regarded as the elements of new latent variables set $\{q(z_{1\_1}), q(z_{1\_2})..q(z_{1\_n/2})\}$ ($(n+1)/2$ variables in total when $n$ is odd, and $n/2$ in total when $n$ is even), then we have below iteration relations:
$$TC_{joint\_2}(z) = TC_{joint\_1}(z) - MU_{joint\_2}(z),$$
$$TC_{joint\_3}(z) = TC_{joint\_2}(z) - MU_{joint\_3}(z),$$
$$....$$
(4)

Finally, there is the following objective:
$$TC(z) = MU_{joint\_1}(z) + MU_{joint\_2}(z) + ... + I(z_{f\_1}; z_{f\_2}),$$
(5)

where $f = int(log_2 n) - 2$. If $f < 1$, then $z_{f\_i} = z_i$, and $n$ is the total dimension of $z$, and $int(c)$ is the rounded-up integer of $c$.

Only two joint distribution variables will be left in the final round of decomposition regardless of the parity of $n$. In objective (5), the remaining joint distribution of the last term is represented by $z_{f\_1}$ and $z_{f\_2}$. $z_{f\_1}$ and $z_{f\_2}$ are two joint distributions composed of distributions selected by the $cp(i,j)$ of the $f$th round. Notice that, only the mutual information $I(z_{f\_1}; z_{f\_2})$ obtained in the last round of decomposition is not based on importance sampling. All $MU_{joint\_x}(z)$ are based on importance sampling. We found that there are also lots of estimates based on importance sampling in the other two decomposition paths. It was empirically demonstrated that importance sampling will limit the usage scenarios of the above decomposition paths. (The experiments and analysis are listed in Supplementary Material A.5)

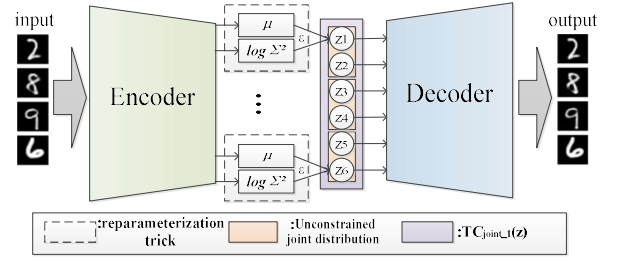

Figure 1. The β−STCVAE structure is almost the same as VAE, as shown in Fig 1. The relationship of disentangling constraints between latent variables becomes more flexible. The number of the latent variable z is 6 in Fig 1. We relieve the local disentangling constraints of the adjacent pair of latent variables and get three joint distributions (Orange area).

### 3.2. New algorithm $\beta$−STCVAE

The real data features $x \in R^N$ are assumed to be divided into two groups: conditionally independent features $v \in R^K$ and conditionally dependent features $v \in R^K$ [16]. The conditional independent features satisfy the independence $log p(v|x) = \sum_K log(v_k|x)$. It is assumed that the real data distribution can be simulated by a parametric neural network $p(x|v,w) = net(v,w)$, using a parametric neural network to find an inferred posterior distribution $q(z|x)$. The minimal complexity disentangling method between latent variables is searched during model training by constraining the total correlation, while the dependent features are loaded in a single latent variable. The disentangling of $\beta$−TCVAE challenges the parameter capacity allocation of neural networks. Assuming the total parameter capacity ($Cz$) of the network consists of two parts: parameter capacity allocated to independent features ($Cv$) and parameter capacity allocated to dependent features ($Cw$), thus $Cz = Cv + Cw$. The parameter capacity can be effectively utilized when it is properly allocated, otherwise, the learning ability of the model will be weakened, wasting the network parameter capacity.

We can constrain $TC_{joint\_x}(z)$ and $MU_{joint\_x}(z)$ separately based on the inference 2 in section 3.1. Intuitively, suppose the constraints of $MU_{joint\_x}(z)$ are omitted, then $C_w$ can be increased while $C_z$ remains unchanged, leading to a reduction in the capacity allocated to the independent distribution features $C_v$. This gives us a method to directly manipulate the parameter capacity allocation in VAE.

According to objective (4), $TC(z)$ can stop decomposition at any round of iteration, and release the disentangling constraints of the first $x$−1 term $MU_{joint\_1}(z) \sim MU_{joint\_x-1}(z)$. The total correlation disentangling penalty terms at any iteration can be written as:

$$TC_{joint\_x}(z)=\begin{cases}D_{KL}(q(z)||\prod_{cp(x-1\_l,x-1\_m)}q(z_{x-1\_l},z_{x-1\_m})) &, n \text{ is even}\\ D_{KL}(q(z)||\prod_{cp(x-1\_l,x-1\_m)}q(z_{x-1\_l},z_{x-1\_m})\cdot q(z_r)) &, n \text{ is odd}\end{cases}$$
(6)

where $n$ is the number of latent variables at current iteration. If $x - 1 \leq 0$, then $z_{x-1\_l} = z_l$, and $z_{x-1\_m} = z_m$.

The $cp()$ selection function in $TC_{joint\_x}(z)$ should be specified, motivated by the decomposition of the total correlation while simplifying the calculation. The $TC_{joint\_x}(z)$ term with penalty coefficient $\beta$ should be used to replace the total correlation term in the $\beta$-TCVAE loss function, then an objective with hyperparameters is developed as the loss function of $\beta$-STCVAE:

$$L_{\beta-STCVAE} := E_{q(z|x)p(x)}[log p(x|z)] - I(z;x) \\ - \beta[TC_{joint}(z)] \\ - \sum_j D_{KL}(q(z_j)||p(z_j)),$$
(7)

Where we take $TC_{joint}(z)$ as $TC_{joint\_x}(z)$'s specific objective, and $TC_{joint}(z) := D_{KL}(q(z)||\prod_{j=1}^{n/i} q(z_{i(j-1)+1}, z_{i(j-1)+2}, ..., z_{ij}))$. The $i$ in $TC_{joint}(z)$ is a factor of the total dimension $n$, named the grouping factor.

Coincidentally, with a completely different decomposition, the loss function of $\beta$-STCVAE is similar to HFVAE. The HFVAE loss function can be written as:

$$L_{HFVAE} := E_{q(z|x)p(x)}[log p(x|z)] - I(z;x) - \beta[TC_{joint}(z)] \\ - \gamma \sum_{j=1}^{\frac{n}{i}} [TC_{joint\_j}(z_{sub})] \\ - \sum_j D_{KL}(q(z_j)||p(z_j)),$$
(8)

Where $TC_{joint\_j}(z_{sub}) := D_{KL}(q(z_{sub})||\prod_{k=i(j-1)+1}^{ij} q(z_k))$, and $q(z_{sub}) := q(z_{i(j-1)+1}, z_{i(j-1)+2}, ..., z_{ij})$.

In the work of HFVAE, the author did not discuss how to select grouping factor $i$ under continuous variables, but mainly studied unsupervised learning of discrete labels. $\beta$-STCVAE discards the disentangling constraints in the sub-variables grouping $\sum_{j=1}^{n/i}[TC_{joint\_j}(z_{sub})]$ compared with HFVAE, gets rid of hyperparameter $\gamma$ that needs to be set training HFVAE. $\beta$-STCVAE becomes $\beta$-TCVAE when $i$ equals 1. All single latent variable dimensions in $\beta$-STCVAE use Gaussian distribution in this paper.

From another perspective, discarding constraint $TC_{joint\_j}(z_{sub})$ makes $\beta$-STCVAE a dynamic prior distribution model. The prior of $\beta$-TCVAE is an isotropic multi-dimensional Gaussian distribution, and the marginal distribution of each dimension is independent of the other.

In $\beta$-STCVAE's prior, although the single dimension's marginal distribution is still the Gaussian distribution, independent constraints between local marginal distributions are relieved. Therefore the prior of latent $z$ is no longer the isotropic multi-dimensional Gaussian distribution, but $n/i$ independent joint distributions (not necessarily multi-dimensional Gaussian distributions) composed of One-dimensional Gaussian distributions.

Intuitively, the grouping factor $i$ should be adjusted based on the model learning scenario. The model's independent prior constraints will be weaker as $i$ increases, and local joint distribution becomes more complex. In this case, $\beta$-STCVAE's encoder can learn more complex $q(z|x)$, which is meaningful for disentangling features with a high degree of complexity, For example, classification tasks of high complexity content. The smaller $i$ is, the stronger the independent constraints of the model prior is. When $i$ equals 1, $\beta$-STCVAE is equivalent to $\beta$-TCVAE, and the prior distribution gradually becomes the multi-dimensional Gaussian distribution with independent marginal distributions. At this point, the model is more suitable for disentangling style details with lower complexity.

## 4. Experiments

### 4.1. Basic Settings

We analyze the feasibility and disentangling performance of $\beta$-STCVAE on five datasets, dSprites [16], celebA [2], shapes3d [18], cars3d [58], and mnist [59].

The experiment's neural network model parameter setting is presented in supplementary materials B.1 as Table1 and Table2

### 4.2. Parameter capacity allocation experiments

We designed the optimal capacity allocation experiments to observe the effect of latent variable grouping factor size on learning ability. The optimal ELBO under different parameter capacities is revealed by varying the latent variable grouping factor. We performed the experiments on five different datasets. The unified experimental settings are as follows:

- The selection list of latent variable dimensions is {6, 8, 10, 12, 14, 16, 18, 20}.
- 20000 iterations for a single training round.
- A single model is repeatedly trained for 20 rounds to ensure accuracy.

Network type selection and parameter capacity list for different datasets refer to Supplementary materials B.1 Table4. We observed the performance of $\beta$-STCVAE with each grouping factor of dimension $n$ (except $n$ itself) at different total parameter capacities. For example, when the latent variable's dimension is 12, the grouping factors' list of experimental is {1, 2, 3, 4, 6}. The β−STCVAE is equivalent to β−TCVAE when grouping factor equals 1. It is necessary to normalize grouping factors to get a unified view of parameter capacity allocation under a different number of latent variables. We take the largest factor $m$ of the grouping factors as the normalization constant, all the selected grouping factors $i$ are recorded as $i/m$ in the experimental results, becoming a coefficient less than or equal to 1 and greater than $1/m$, called the latent variable $z$'s grouping coefficient. When $z$'s dimension is 12, the experimental grouping coefficient list is {1/6, 2/6, 3/6, 4/6, 6/6}. The grouping coefficient represents the parameter capacity allocation of the model. The larger the grouping coefficient, the stronger the learning ability to independent features. Conversely, the smaller the grouping coefficient, the stronger the learning ability to independent features. We recorded the grouping coefficient of the best ELBO under each parameter capacity and plotted the average best ELBO trajectory under different parameter capacities.

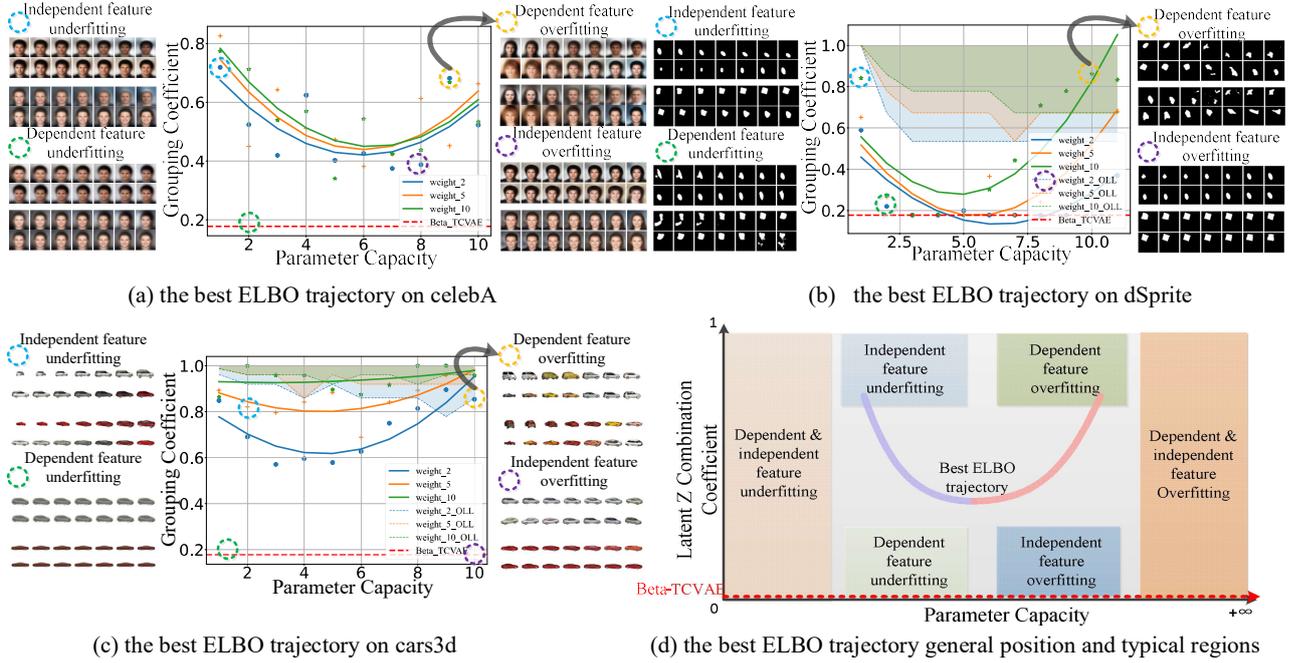

Figure 2. The Fig. 2(a),(b),(c) plot the best ELBO trajectory of the model under different parameter capacities on datasets. The digit on the parameter capacity-axis corresponds to the element index in the parameter capacity list in Supplementary materials B. Table 4 . The red dotted line below the figure represents β−TCVAE when the grouping factor is 1. Grouping coefficients corresponding to β−TCVAE in different dimensions are inconsistent, so we take the average value 0.178 as the drawing reference (The red dotted line below the Figs). Except for the dSprites training set, β−TCVAE is not on the ideal ELBO trajectory, regardless of how parameters and disentangling coefficients are altered in the datasets. The left and right of each Figs are images generated by latent traversing at 4 distinct regions (sampling in the circular dotted area). The lower Fig. 2(d) shows a general position of the optimal ELBO trajectory and indicates the approximate generation regions for six kinds of typical samples. In dSprites, cars3d, and shape3d, we marked the occurrence region (_OLL color area in the Fig. 2(b) and Fig. 2(c)) of omniscient latent variable when $\epsilon=0.001$ and $\delta=0.01$ under various disentangling strengths.

We trained 100000+ β−STCVAE models on five datasets in total (see the zoom-in experiments' results for mnist and shape3D in Supplementary materials C). Leverage a binary function curve to fit these optimal ELBO points. Intriguingly, the V-shaped pattern was found on every dataset. Fig. 2 shows that β−STCVAE with a reasonable allocation of parameter capacity can always obtain better ELBO than β−TCVAE (red dotted line) except for some situations in the dSprites dataset, regardless of disentangling strength and parameter capacities.

**Interpretation of the V-shaped optimal ELBO trajectory.** Learning dependent data features require more than the parameter capacity assigned to a single latent variable when the total parameter capacity is low, thus optimal ELBO tends to release local disentangling constraints to learn more data features. Then, the parameter capacity assigned to a single latent variable dimension rises along with the gradually increasing total parameter capacity, which explains that the grouping coefficient of the best ELBO decreases as total parameter capacity gradually increases (purple line area of Fig. 2(d)). When a critical point is reached, because the features learned by the model parameter capacity are complex enough, a matching complex latent variable distribution is required to accommodate the features. Therefore, the one-dimensional Gaussian distributions are combined to create a more complex prior distribution, causing the grouping coefficient under the best ELBO to rise again (pink line area of Fig. 2(d)). Additionally, we discovered that as the disentangling strength enhances, the optimal ELBO trajectory moves up. This phenomenon implies that the model allocates more parameter capacity to learn dependent features as a balancing mechanism to counteract the high disentangling strength.

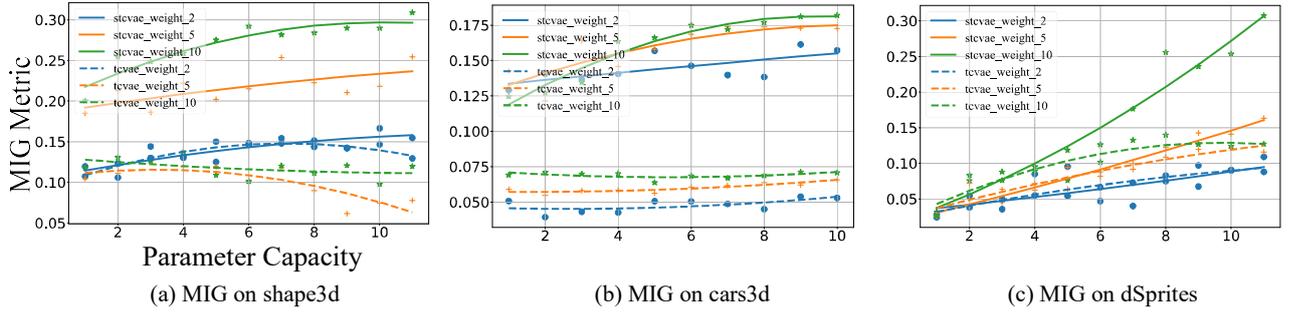

(a) MIG on shape3d  (b) MIG on cars3d  (c) MIG on dSprites

Figure 3. The left, middle, and right are the average MIG of the two models under different parameter capacities on shape3d, cars3d, and dSprites datasets.

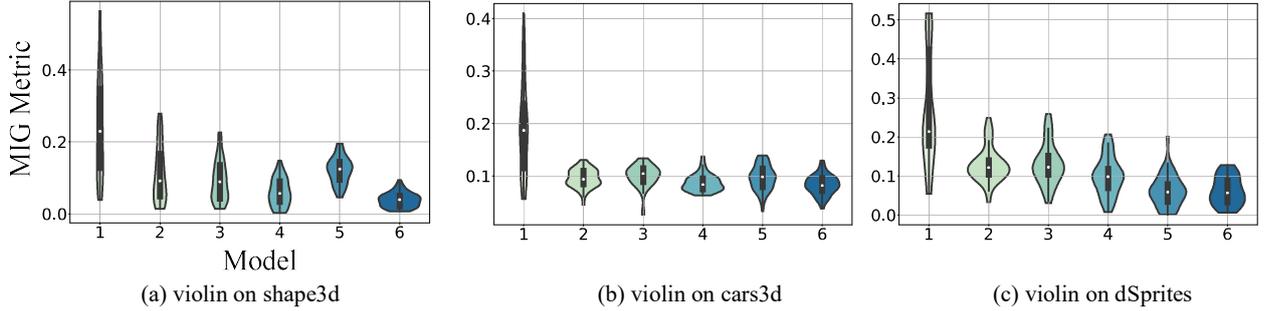

(a) violin on shape3d  (b) violin on cars3d  (c) violin on dSprites

Figure 4. The Fig. 4(a), Fig. 4(b) and Fig. 7(c) are the MIG-based disentangling metric violin graphs of different VAE models on shape3d, cars3d, and dSprites datasets. Models are abbreviated (1=StcVae_8_2, 2=TcVae_16, 3=TcVae_8, 4=BetaVae_8, 5=FactorVae_8, 6=DIP-I_8). Abbreviated writing method: Model name_latent variable dimension_Grouping factor. For example, Stcvae_8_2 represents a Stcvae model with 8 latent variables when the total number of latent dimensions is 16, and the grouping factor is 2. TcVae_8 represents a TcVae model with 8 latent dimensions.

**Typical generated sample characteristics in different regions.** After the generation of numerous samples from various datasets, we can summarize the following laws (Refer to Fig. 2(d) and Supplementary materials C):

- Dependent feature overfitting on the Upper right of Fig. 2(d): Caused by $Cw$ being too high, typically when traversing a single latent variable. The generated samples' features change significantly and contain various features that cannot be assigned to a specific attribute or category.
- Dependent feature underfitting on the Lower left of Fig. 2(d): Caused by $Cw$ being too low, typically when traversing a single latent variable. The generated samples are reconstructions of input samples with low or corrupt quality.
- Independent features overfitting on the Lower right of Fig. 2(d): Caused by $Cv$ being too high. The typical performance is that two groups of generated samples are obtained when traversing two latent variables. Only slight differences in details can be regarded as entangling between the transformation methods of the two sample groups.
- Independent features Underfitting on the Upper left of Fig. 2(d): Caused by $Cv$ being too low. The typical performance is that two groups of generated samples are obtained when traversing two latent variables. There are exaggerated differences between the transformation methods of the two sample groups, far beyond the subjective concept of disentangling.

Correctly assigning the grouping factor remains a challenge, but the regional characteristics of generated samples can be referred to manipulate the degree of disentangling. When classifying distinct types of samples, the parameters should be near the independent feature underfitting region in the upper left corner. Conversely, when decoupling the styles within a category, make the model parameters in the lower right independent feature overfitting region. This also explains the experimental results in HFVAE[33]. We speculate: The latent traversal of HFVAE can better decouple the subtle data style than $\beta$VAE because HFVAE somehow appears in the lower right independent feature overfitting region by adjusting the two-level disentangling strengths.

**Omniscient latent variable.** In the experiments, we discovered that when the latent variable grouping factor reaches a certain value threshold, the latent variable will become overloaded with data features if the parameter capacity is high enough. The phenomenon is manifested as the dependent features overfitting during latent traversal on the latent variable, called the omniscient latent variable. So, we try to locate the occurrence range of this phenomenon in cars3D, dSprites, and shape3D datasets with ground true factor labels. We found that when calculating the latent variable marginal and conditional entropy, the omniscient latent variable behaves as a variable whose information entropy is 0. Meaning the variable cannot provide any information because it attempts to contain the characteristic states of all samples. During training, it was also discovered that the omniscient latent variable phenomenon only appears on one single latent variable in most training.

First, we define the omniscient latent variable as the probability that any dimension's information entropy in latent variables $z$ of all trained models satisfies:

$$P[E_{q(z_i)}(-log q(z_i)) < \epsilon] \geq 1 - \delta, \tag{9}$$

under a given parameter capacity. Fig. 2(b) shows that the omniscient latent variable appears at the top right in dSprites, while it is more concentrated at the top in the other two datasets (See experiments on shape3D in Supplementary materials C). The reason for the difference is that the model was trained using MLP network on the dSprites dataset while using Conv network on the other two datasets. The overall parameter capacity of Conv is greater than MLP because the convolutional layer cannot alter parameter capacity, as designed in Supplementary materials B.1. Therefore, it has a certain "basic" parameter capacity in Conv network that makes the parameter space shift to the right as a whole. The omniscient latent variable appears in the upper right side of the figure, indicating this phenomenon occurs in the region of dependent features overfitting, which is in line with our definition in typical generated sample characteristics.

The MIG metric method uses the first-highest mutual information minus the second-highest mutual information to achieve a penalty for multiple variables containing the same features. However, if there are only two latent variables and one of them is an omniscient latent variable, then the second highest mutual information will always be 0, resulting in MIG distortion. Therefore, the $\beta$−STCVAE with only two latent variables should be avoided when calculating MIG in the region where omniscient latent variables occur.

### 4.3. Disentangling capability evaluation

The MIG disentangling measures of $\beta$−TCVAE and $\beta$−STCVAE under different parameter capacities and latent variable grouping coefficients are tested on the cars3D, dSprites, and shape3D, in Fig. 3. Model $\beta$−STCVAE takes the average value of MIG under all grouping coefficients except 1 (to avoid the MIG distortion caused by the omniscient latent variable).

Set the model referring to 4.4.1, and take the training model's average MIG. As shown in Fig. 3, the disentangling ability of $\beta$−STCVAE enhances obviously with the increase of parameter capacity. In contrast, the disentangling ability of $\beta$−TCVAE enhances gently (some even decline). We think the main reason is that under different parameter capacities, $\beta$−STCVAE can always find the appropriate prior distribution of latent variables to accommodate the data features learned by model, while $\beta$−TCVAE cannot adjust the prior distribution of latent variables, which limits the expressiveness of disentangling.

The MIG disentangling metrics of $\beta$−STCVAE, $\beta$−TCVAE, $\beta$VAE, FactorVAE, and DIP-VAE-I are tested on 3 datasets under a fixed number of latent variables, as shown in Fig. 4. Setting model hyperparameter referring Supplementary materials B.1 Table3. The dSprites dataset parameter capacity list is {3000, 4000, 5000}; cars3d parameter capacity list is {800, 900, 1000}; and shape3d parameter capacity list is {800, 1000, 1200}. The disentangling performances of $\beta$−STCVAE on these three datasets are significantly better than other VAEs.

## 5. Conclusion

Most previous research in academia has concentrated on how to make the disentangling capabilities of VAE more human-comprehensible. But, there are a thousand Hamlets in a thousand people's eyes. Even for humanity, the feature disentangling under different subjective concepts varies. We should pay more attention to the controllability of the unsupervised disentangling capability, rather than being satisfied that it meets the evaluation criteria from a certain perspective

A dynamic disentangled prior distribution is created in this research by decomposing the independence constraints in multi-dimensional Gaussian distribution, allowing VAE to be utilized in more complex disentangling scenarios. The proposed approach creates a Macro and micro controllable unsupervised disentangling algorithm, which is more in line with human understanding. In the experiment, the best ELBO trajectory in parameter space is visualized, as are the characteristics of generated samples in corresponding regions, providing some suggestions for future work. It also demonstrates that the new method can provide more flexible disentangling performance.

However, unified disentangling capability metrics for complex data features are still lacking. The data properties in the real world are unquestionably more complicated, but the evaluations of new methods are only being conducted on toy datasets. In the future, we propose to describe the difference in disentangling degree produced by various subjective perspectives.

# References

[1] D. P. Kingma and M. Welling, "Auto-encoding variational bayes," arXiv preprint arXiv:1312.6114, 2013.

[2] R. T. Chen, X. Li, R. B. Grosse, and D. K. Duvenaud, "Isolating sources of disentanglement in variational autoencoders," Advances in neural information processing systems, vol. 31, 2018.

[3] G. Lee and H. Li, "Modeling code-switch languages using bilingual parallel corpus," in Proceedings of the 58th Annual Meeting of the Association for Computational Linguistics, 2020, pp. 860-870.

[4] X. Chen, "Simulation of English speech emotion recognition based on transfer learning and CNN neural network," Journal of Intelligent & Fuzzy Systems, vol. 40, no. 2, pp. 2349-2360, 2021.

[5] Y. Lü, H. Lin, P. Wu, and Y. Chen, "Feature compensation based on independent noise estimation for robust speech recognition," EURASIP Journal on Audio, Speech, and Music Processing, vol. 2021, no. 1, pp. 1-9, 2021.

[6] Y. Shi, X. Yu, K. Sohn, M. Chandraker, and A. K. Jain, "Towards universal representation learning for deep face recognition," in Proceedings of the IEEE/CVF Conference on Computer Vision and Pattern Recognition, 2020, pp. 6817-6826.

[7] T. Ni, X. Gu, C. Zhang, W. Wang, and Y. Fan, "Multi-task deep metric learning with boundary discriminative information for cross-age face verification," Journal of Grid Computing, vol. 18, no. 2, pp. 197-210, 2020.

[8] X. Shi, C. Yang, X. Xia, and X. Chai, "Deep cross-species feature learning for animal face recognition via residual interspecies equivariant network," in European Conference on Computer Vision, 2020: Springer, pp. 667-682.

[9] J. Chen, B. Lei, Q. Song, H. Ying, D. Z. Chen, and J. Wu, "A hierarchical graph network for 3d object detection on point clouds," in Proceedings of the IEEE/CVF conference on computer vision and pattern recognition, 2020, pp. 392-401.

[10] A. S. Brown, S. Hornstein, and A. Memon, "Tracking conversational repetition: An evaluation of target monitoring ability," Applied Cognitive Psychology: The Official Journal of the Society for Applied Research in Memory and Cognition, vol. 20, no. 1, pp. 85-95, 2006.

[11] Z. Xu, E. Hrustic, and D. Vivet, "Centernet heatmap propagation for real-time video object detection," in European conference on computer vision, 2020: Springer, pp. 220-234.

[12] D. Zhang, H. Tian, and J. Han, "Few-cost salient object detection with adversarial-paced learning," Advances in Neural Information Processing Systems, vol. 33, pp. 12236-12247, 2020.

[13] A. Torfi, R. A. Shirvani, Y. Keneshloo, N. Tavaf, and E. A. Fox, "Natural language processing advancements by deep learning: A survey," arXiv preprint arXiv:2003.01200, 2020.

[14] S. Stoll, N. C. Camgoz, S. Hadfield, and R. Bowden, "Text2Sign: towards sign language production using neural machine translation and generative adversarial networks," International Journal of Computer Vision, vol. 128, no. 4, pp. 891-908, 2020.

[15] P. He, X. Liu, J. Gao, and W. Chen, "Deberta: Decoding-enhanced bert with disentangled attention," arXiv preprint arXiv:2006.03654, 2020.

[16] I. Higgins et al., "beta-vae: Learning basic visual concepts with a constrained variational framework," 2016.

[17] A. Kumar, P. Sattigeri, and A. Balakrishnan, "Variational inference of disentangled latent concepts from unlabeled observations," arXiv preprint arXiv:1711.00848, 2017.

[18] H. Kim and A. Mnih, "Disentangling by factorising," in International Conference on Machine Learning, 2018: PMLR, pp. 2649-2658.

[19] M. Kim, Y. Wang, P. Sahu, and V. Pavlovic, "Relevance factor VAE: Learning and identifying disentangled factors," arXiv preprint arXiv:1902.01568, 2019.

[20] F. Locatello et al., "Challenging common assumptions in the unsupervised learning of disentangled representations," in international conference on machine learning, 2019: PMLR, pp. 4114-4124.

[21] S. Parekh, S. Essid, A. Ozerov, N. Q. Duong, P. Pérez, and G. Richard, "Weakly supervised representation learning for audio-visual scene analysis," IEEE/ACM Transactions on Audio, Speech, and Language Processing, vol. 28, pp. 416-428, 2019.

[22] F. Locatello, B. Poole, G. Rätsch, B. Schölkopf, O. Bachem, and M. Tschannen, "Weakly-supervised disentanglement without compromises," in International Conference on Machine Learning, 2020: PMLR, pp. 6348-6359.

[23] J. Antoran and A. Miguel, "Disentangling and learning robust representations with natural clustering," in 2019 18th IEEE International Conference On Machine Learning And Applications (ICMLA), 2019: IEEE, pp. 694-699.

[24] L. Bonheme and M. Grzes, "Be More Active! Understanding the Differences between Mean and Sampled Representations of Variational Autoencoders," arXiv preprint arXiv:2109.12679, 2021.

[25] Y. Burda, R. Grosse, and R. Salakhutdinov, "Importance weighted autoencoders," arXiv preprint arXiv:1509.00519, 2015.

[26] A. Vahdat and J. Kautz, "NVAE: A deep hierarchical variational autoencoder," Advances in Neural Information Processing Systems, vol. 33, pp. 19667-19679, 2020.

[27] A. Shekhovtsov, D. Schlesinger, and B. Flach, "VAE Approximation Error: ELBO and Exponential Families," in International Conference on Learning Representations, 2021.

[28] E. Dupont, "Learning disentangled joint continuous and discrete representations," Advances in Neural Information Processing Systems, vol. 31, 2018.

[29] C. P. Burgess et al., "Understanding disentangling in $\beta$-VAE," arXiv preprint arXiv:1804.03599, 2018.

[30] H. Shao et al., "Controlvae: Controllable variational autoencoder," in International Conference on Machine Learning, 2020: PMLR, pp. 8655-8664.

[31] S. Zhao, J. Song, and S. Ermon, "Learning hierarchical features from deep generative models," in International Conference on Machine Learning, 2017: PMLR, pp. 4091-4099.

[32] M. Willetts, S. Roberts, and C. Holmes, "Disentangling to cluster: Gaussian mixture variational Ladder autoencoders," arXiv preprint arXiv:1909.11501, 2019.

[33] B. Esmaeili et al., "Structured disentangled representations," in The 22nd International Conference on Artificial Intelligence and Statistics, 2019: PMLR, pp. 2525-2534.

[34] D. Bouchacourt, R. Tomioka, and S. Nowozin, "Multi-level variational autoencoder: Learning disentangled

representations from grouped observations," in Proceedings of the AAAI Conference on Artificial Intelligence, 2018, vol. 32, no. 1.
[35] A. Szabo, Q. Hu, T. Portenier, M. Zwicker, and P. Favaro, "Understanding degeneracies and ambiguities in attribute transfer," in Proceedings of the European Conference on Computer Vision (ECCV), 2018, pp. 700-714.
[36] Y. Ge, S. Abu-El-Haija, G. Xin, and L. Itti, "Zero-shot synthesis with group-supervised learning," arXiv preprint arXiv:2009.06586, 2020.
[37] E. H. Sanchez, M. Serrurier, and M. Ortner, "Learning disentangled representations via mutual information estimation," in European Conference on Computer Vision, 2020: Springer, pp. 205-221.
[38] P. Esser, J. Haux, and B. Ommer, "Unsupervised robust disentangling of latent characteristics for image synthesis," in Proceedings of the IEEE/CVF International Conference on Computer Vision, 2019, pp. 2699-2709.
[39] D. Lorenz, L. Bereska, T. Milbich, and B. Ommer, "Unsupervised part-based disentangling of object shape and appearance," in Proceedings of the IEEE/CVF Conference on Computer Vision and Pattern Recognition, 2019, pp. 10955-10964.
[40] K. Greff, A. Rasmus, M. Berglund, T. Hao, H. Valpola, and J. Schmidhuber, "Tagger: Deep unsupervised perceptual grouping," Advances in Neural Information Processing Systems, vol. 29, 2016.
[41] Y. Li, K. K. Singh, U. Ojha, and Y. J. Lee, "Mixnmatch: Multifactor disentanglement and encoding for conditional image generation," in Proceedings of the IEEE/CVF conference on computer vision and pattern recognition, 2020, pp. 8039-8048.
[42] C. Cremer, X. Li, and D. Duvenaud, "Inference suboptimality in variational autoencoders," in International Conference on Machine Learning, 2018: PMLR, pp. 1078-1086.
[43] D. Hjelm, R. R. Salakhutdinov, K. Cho, N. Jojic, V. Calhoun, and J. Chung, "Iterative refinement of the approximate posterior for directed belief networks," Advances in neural information processing systems, vol. 29, 2016.
[44] Y. Kim, S. Wiseman, A. Miller, D. Sontag, and A. Rush, "Semi-amortized variational autoencoders," in International Conference on Machine Learning, 2018: PMLR, pp. 2678-2687.
[45] J. He, D. Spokoyny, G. Neubig, and T. Berg-Kirkpatrick, "Lagging inference networks and posterior collapse in variational autoencoders," arXiv preprint arXiv:1901.05534, 2019.
[46] B. Dai, Z. Wang, and D. Wipf, "The usual suspects? Reassessing blame for VAE posterior collapse," in International Conference on Machine Learning, 2020: PMLR, pp. 2313-2322.
[47] S. Nowozin, "Debiasing evidence approximations: On importance-weighted autoencoders and jackknife variational inference," in International conference on learning representations, 2018.
[48] D. Rezende and S. Mohamed, "Variational inference with normalizing flows," in International conference on machine learning, 2015: PMLR, pp. 1530-1538.
[49] R. Ranganath, D. Tran, and D. Blei, "Hierarchical variational models," in International conference on machine learning, 2016: PMLR, pp. 324-333.
[50] B. Jiang, C. Ye, and J. S. Liu, "Nonparametric k-sample tests via dynamic slicing," Journal of the American Statistical Association, vol. 110, no. 510, pp. 642-653, 2015.
[51] C. Granger and J. L. Lin, "Using the mutual information coefficient to identify lags in nonlinear models," Journal of time series analysis, vol. 15, no. 4, pp. 371-384, 1994.
[52] P. Cheng et al., "Improving disentangled text representation learning with information-theoretic guidance," arXiv preprint arXiv:2006.00693, 2020.
[53] R. D. Hjelm et al., "Learning deep representations by mutual information estimation and maximization," arXiv preprint arXiv:1808.06670, 2018.
[54] D. J. Zea, D. Anfossi, M. Nielsen, and C. Marino-Buslje, "MIToS. jl: mutual information tools for protein sequence analysis in the Julia language," Bioinformatics, vol. 33, no. 4, pp. 564-565, 2017.
[55] A. Lachmann, F. M. Giorgi, G. Lopez, and A. Califano, "ARACNe-AP: gene network reverse engineering through adaptive partitioning inference of mutual information," Bioinformatics, vol. 32, no. 14, pp. 2233-2235, 2016.
[56] J.-F. Cardoso, "Dependence, correlation and gaussianity in independent component analysis," The Journal of Machine Learning Research, vol. 4, pp. 1177-1203, 2003.
[57] P. Cheng, W. Hao, and L. Carin, "Estimating total correlation with mutual information bounds," arXiv preprint arXiv:2011.04794, 2020.
[58] S. E. Reed, Y. Zhang, Y. Zhang, and H. Lee, "Deep visual analogy-making," Advances in neural information processing systems, vol. 28, 2015.
[59] L. Deng, "The mnist database of handwritten digit images for machine learning research [best of the web]," IEEE signal processing magazine, vol. 29, no. 6, pp. 141-142, 2012.